\documentclass[10pt,twocolumn,letterpaper]{article}

\usepackage[pagenumbers]{cvpr} 

\usepackage{graphicx}
\usepackage{amsmath,amssymb,amsthm,mathtools}
\usepackage{bm}
\usepackage{booktabs}
\usepackage{multirow}
\usepackage{makecell}
\usepackage{tabularx}
\usepackage{array}
\usepackage{xspace}
\usepackage{microtype}
\usepackage{enumitem}
\usepackage{pifont}
\usepackage{placeins}
\usepackage{etoolbox}
\usepackage{cuted}
\usepackage{caption}
\usepackage{tikz}
\usetikzlibrary{arrows.meta,positioning,calc}
\usepackage{pgfplots}
\pgfplotsset{compat=1.18}

\newcommand{\ours}{LangStreet\xspace}
\newcommand{\kitti}{KITTI-360\xspace}
\newcommand{\waymo}{Waymo\xspace}

\newcommand{\vkitti}{vKITTI2\xspace}

\newcommand{\psnr}{PSNR\xspace}

\newcommand{\up}{$\uparrow$}
\newcommand{\down}{$\downarrow$}

\newcommand{\Sph}{\mathbb{S}}

\newcommand{\fulltablewidth}{0.95\textwidth}

\newcommand{\mainwidefont}{\fontsize{7.35}{8.35}\selectfont}
\newcommand{\mainsinglefont}{\fontsize{7.15}{8.10}\selectfont}

\definecolor{lsnavy}{HTML}{24324A}
\definecolor{lsblue}{HTML}{3F6FB5}
\definecolor{lsteal}{HTML}{2E8B83}
\definecolor{lsorange}{HTML}{D98B3A}
\definecolor{lsred}{HTML}{B84A4A}
\definecolor{lsink}{HTML}{263238}
\definecolor{lsmuted}{HTML}{667085}
\definecolor{lsline}{HTML}{D7DCE3}
\definecolor{lspanel}{HTML}{F7F9FC}

\makeatletter
\newcommand{\rowinput}[1]{\input{#1} }
\makeatother
\AtBeginEnvironment{tabular}{\let\input\rowinput}

\microtypesetup{protrusion=true,expansion=true}
\AtBeginDocument{\setlength{\parfillskip}{0pt plus .65\columnwidth}}
\setlist[itemize]{leftmargin=*,topsep=2pt,itemsep=1pt,parsep=0pt}

\definecolor{cvprblue}{rgb}{0.21,0.49,0.74}
\usepackage[pagebackref,breaklinks,colorlinks,allcolors=cvprblue]{hyperref}

\def\paperID{101}
\def\confName{3DV\xspace}
\def\confYear{2027\xspace}

\title{LangStreet: Persistent Language Fields for Anchor-Decoded Street Gaussians}
\author{
Runyi Yang$^{1}$ \quad Deheng Zhang$^{1}$ \quad Xiaoye Wang$^{1}$ \quad Mengjiao Ma$^{1}$ \quad Lei Sun$^{1}$\\
Kanzhi Wu$^{2}$ \quad Ajad Chhatkuli$^{1}$ \quad Luc Van Gool$^{1}$ \quad Danda Pani Paudel$^{1}$\\[4pt]
{\normalsize $^{1}$INSAIT, Sofia University ``St. Kliment Ohridski''\quad $^{2}$vivo Mobile Communication Co., Ltd., Shenzhen, China}
}

\begin{document}
\maketitle
\begin{strip}
  \centering

  \includegraphics[
    width=0.98\textwidth
  ]{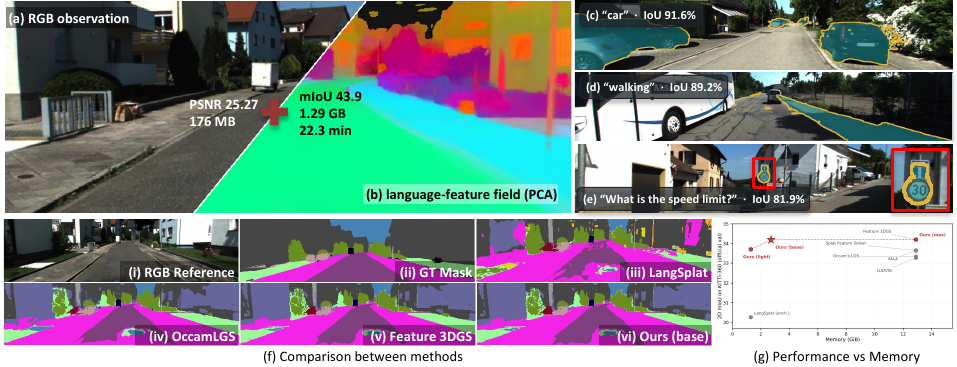}

  \begin{minipage}{0.96\textwidth}
    \captionsetup{
      justification=justified,
      singlelinecheck=false
    }
\captionof{figure}{
  Using posed RGB images and fixed pretrained priors, \ours assigns language
  to persistent anchors and slots without scene labels. Panels (a) and (b)
  show an RGB observation and the language field using the first three
  principal components. Panels (c)--(e) illustrate object, affordance, and
  scene-level text queries. Panel (f) compares semantic predictions across
  methods. Panel (g) summarizes the accuracy--storage trade-off:
  \ours~(base) nearly matches the full-slot field while using substantially
  less feature storage.
}
    \label{fig:teaser}
  \end{minipage}
\end{strip}

\begin{abstract}
Language Gaussian fields implicitly assume that the primitive carrying semantics remains identifiable across views. This assumption breaks in scalable anchor-decoded representations, where persistent anchors generate view-conditioned child Gaussians whose geometry and appearance vary with the camera. We introduce \textbf{\ours}, a persistent language field for such structured Gaussian scenes. Our key idea is \emph{semantic ownership}: transient children route observations, while persistent decoder slots and their parent anchors own the language field. We use alpha-compositing responsibilities to accumulate additive directional evidence at slots; these statistics marginalize exactly to anchors. We then complete weakly supported slots with anchor-aligned evidence while preserving the anchor direction, and represent slot detail through low-rank residuals in anchor-relative semantic coordinates. Our primary model, \ours~(base), stores anchor features together with compact slot residuals. \ours~(light) retains only anchor features, whereas \ours~(max) stores the full-dimensional completed slot features explicitly. Without scene-specific semantic optimization, \ours~(base) nearly matches \ours~(max) across \kitti, \vkitti, and \waymo. On \kitti, it achieves $34.19$ 2D mIoU with a $2.72$\,GiB effective feature footprint, compared with $34.20$ mIoU and $12.90$\,GiB for \ours~(max). The same accuracy--storage trend holds on \vkitti and \waymo. These results show that language fields on view-conditioned splats require persistent semantic ownership, conserved evidence, and a hierarchy that balances stability, detail, and representation cost. Our code, checkpoints, and benchmark suite will be publicly available.
\end{abstract}
\vspace{-2em}
\section{Introduction}
\label{sec:intro}

Street-scale Gaussian representations reconstruct long urban trajectories with
high visual fidelity and interactive rendering
\cite{yan2024street,zhou2024drivinggaussian,zhou2024hugs,chen2025omnire,
liu2024citygaussian,kerbl2024hierarchical}. They are increasingly useful for
autonomous driving, simulation, map maintenance, and urban digital twins, yet
their contents remain difficult to access semantically. Language fields address
this limitation by embedding vision--language features in 3D, enabling textual
queries over scenes
\cite{kerr2023lerf,qin2024langsplat,shi2024legaussians,
zhou2024feature3dgs}. As illustrated in Fig.~\ref{fig:teaser}, our goal is to
turn a street reconstruction into a persistent, language-queryable spatial
asset.

\begin{figure}[t]
  \centering
  \includegraphics[width=\columnwidth]{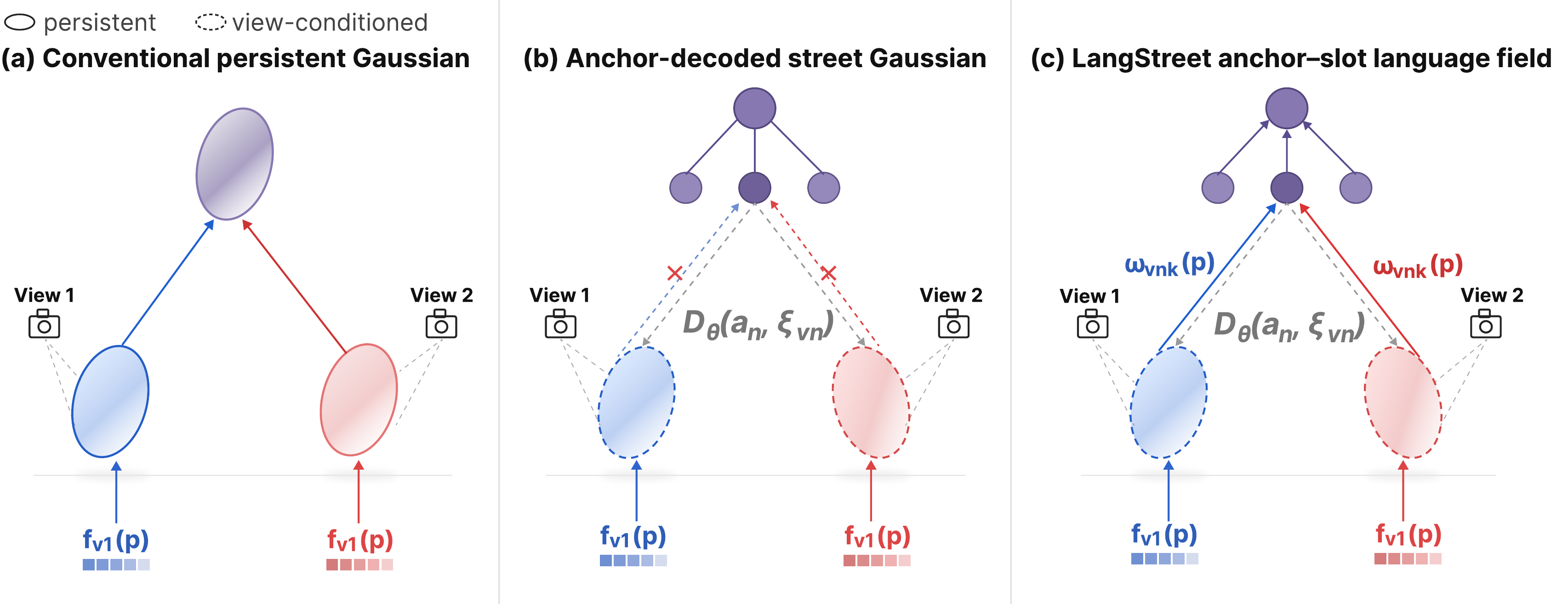}
  \vspace{-1em}
    \caption{\textbf{Semantic ownership in view-conditioned splats.}
    (a) Persistent Gaussians can own language directly.
    (b) Anchor decoders generate view-conditioned children without persistent
    addresses.
    (c) \ours uses children to route observations to persistent slots, whose
    evidence aggregates to anchors.}
  \label{fig:anchor_slot_ownership}
\end{figure}

Making such an asset persistent requires deciding where language fields live.
Conventional language Gaussian fields attach semantics to stored Gaussians
whose indices remain valid across views. Anchor-decoded models break this
correspondence: they store persistent anchors and indexed decoder slots, but
instantiate child Gaussians whose geometry and activity vary with the camera
\cite{lu2024scaffold,ren2025octree}. As shown in
Fig.~\ref{fig:anchor_slot_ownership}, a child can route an observation through
the renderer, but cannot own persistent semantics. We treat the slot as the
finest persistent address and its anchor as a coarser parent. A slot need not
possess semantics beforehand; its semantic content is induced by observations
repeatedly routed to the same address.

This ownership structure creates two coupled challenges. A slot receives
evidence only when its child is active, visible, and contributes to a rendered
pixel, leaving many slots weakly supported or entirely unobserved. Meanwhile,
storing a full-dimensional vision--language feature at every slot repeatedly
encodes content shared within the same anchor. A scalable language field must
therefore preserve fine slot-level detail, recover unsupported slots, and
exploit anchor-level structure for compactness.

We introduce \textbf{\ours}, an anchor--slot language field built around
semantic ownership. As shown in Fig.~\ref{fig:anchor_slot_ownership}(c),
view-conditioned children route 2D observations to persistent slots through
their alpha-compositing responsibilities. Rather than storing only normalized
slot embeddings, we retain their additive directional evidence; summing these
sufficient statistics recovers the parent-anchor evidence exactly. The anchor
then supplies aligned evidence to weakly supported slots while preserving the
coarse anchor direction. Finally, we treat each anchor as a local semantic
origin and encode only low-rank slot residuals, retaining sub-anchor detail
without storing a full-dimensional feature at every slot.

This construction yields three model variants, illustrated in
Fig.~\ref{fig:teaser}. Our primary configuration, \ours~(base), stores one
full-dimensional feature per anchor together with compact anchor-relative
codes for its slots. \ours~(light) retains only the anchor field, while
\ours~(max) stores the completed slot features explicitly as a full-slot
reference. All three are derived from the same renderer-routed evidence without
scene-specific semantic optimization. Because they store vision--language
embeddings rather than fixed class logits, the fields can be scored against new
text queries without rebuilding the 3D representation.

We evaluate \ours under a controlled estimator protocol in which all methods
share the same anchor-decoded geometry, construction views, 2D observations,
prompts, renderer, and evaluator. The compared rows are therefore matched
adaptations of their estimation principles rather than full end-to-end
reproductions. \kitti serves as our main real-street benchmark, \waymo is
evaluated independently, and \vkitti isolates sparse slot evidence. As
previewed in Fig.~\ref{fig:teaser}, \ours~(base) retains nearly all of the
accuracy of \ours~(max) while using substantially less feature storage on
\kitti and \waymo. On \vkitti, hierarchical completion primarily improves
sparse slot-level 3D queries rather than rendered 2D accuracy.

Our contributions are as follows:
\label{sec:intro_contributions}
\begin{itemize}
\vspace{-0.3em}
  \item \textbf{Semantic ownership for view-conditioned splats.}
  We separate transient child Gaussians that route observations from
  persistent slots and anchors that own language.

  \item \textbf{An evidence-conserving hierarchy.}
  Compositing responsibilities accumulate additive slot evidence that
  marginalizes exactly to anchors. Anchor-aligned completion preserves the
  anchor direction, while anchor-relative residuals yield the compact
  \ours~(base) representation.

  \item \textbf{A strong accuracy--storage trade-off at street scale.}
  Across three datasets, \ours~(base) closely matches dense full-slot accuracy
  at a substantially smaller feature footprint. Controlled analysis further
  separates the effects of semantic carrier, completion, coding, and readout.
\end{itemize}
\section{Related Work}
\label{sec:related}

\paragraph{Structured urban Gaussian representations.}
Street-scale Gaussian models scale through object decomposition, spatial
partitioning, and levels of detail
\cite{yan2024street,zhou2024drivinggaussian,zhou2024hugs,chen2025omnire,
liu2024citygaussian,kerbl2024hierarchical,jiang2025horizongs}; SUNDAE instead
compresses them by spectral graph pruning and neural compensation
\cite{yang2024sundae}. Scaffold-GS and Octree-GS decode view-conditioned
children from persistent anchors and slots \cite{lu2024scaffold,ren2025octree},
while Gaussian-JEPA learns reusable embeddings of Gaussian assets
\cite{ren2026gaussian}. These works target reconstruction, compression, or
representation learning; \ours asks which native persistent address should own
language.

\vspace{-1em}
\paragraph{Language fields and compact semantics.}
LERF and OpenNeRF lift image features into radiance fields, while OpenMask3D
aggregates multi-view CLIP evidence for 3D instances
\cite{kerr2023lerf,engelmann2024opennerf,takmaz2023openmask3d}. LangSplat,
LEGaussians, Feature 3DGS, and OpenGaussian attach semantics to explicit
Gaussians, and 4D LangSplat extends this setting to dynamics
\cite{qin2024langsplat,shi2024legaussians,zhou2024feature3dgs,
wu2024opengaussian,li2025fourdlanguagesplat}. GARField learns a separately
optimized hierarchical affinity field \cite{kim2024garfield}. Training-free
estimators lift frozen observations by rendering, diffusion, solving, or
visibility-aware aggregation
\cite{cheng2025occam,marrie2025ludvigofficial,xiong2026sfs,wang2025vala}.
SceneSplat, Chorus, SceneSplat++, and LangFlash study transferable,
generalizable, or feed-forward language splatting
\cite{li2025scenesplat,li2025chorus,ma2025scenesplatpp,liu2026langflash}.
Compact and query-time methods reduce dense storage
\cite{li2025langsplatv2,bang2026lightsplat,budimir2026scoup,jia2026cosag,
lee2025cf3,zhu2026querygaussian}. LangSplatV2 uses a global dictionary with
sparse per-Gaussian coefficients; \ours instead codes slot residuals in local
anchor-relative coordinates. Prior methods generally use persistent Gaussians
or separate semantic structures, whereas ours follows the decoder's native
anchor--slot hierarchy.

\vspace{-1em}
\paragraph{Semantic street scenes.}
MARS decomposes driving scenes into persistent background and foreground
radiance fields, supporting controllable appearance and trajectories together
with RGB, depth, and semantic rendering \cite{wu2023mars}. City-scale continual
mapping couples environment- and instance-level fields
\cite{shi2024cityscale}. Recent outdoor language-Gaussian systems support
open-vocabulary understanding or generation
\cite{szilagyi2025slag,li2026dilegs,yan2026outlangsplat,
deng2026gaussiandwm}, while Query3D interprets questions over an existing field
\cite{chahe2025query3d}. They do not address ownership when transient children
are decoded from persistent anchors and slots.

\section{Preliminaries and Problem Setup}
\label{sec:preliminaries}

\paragraph{Persistent Gaussian addresses.}
A conventional Gaussian scene stores an indexed set of primitives
$\mathcal G=\{g_j\}_{j=1}^{M}$, where
$g_j=(\boldsymbol\mu_j,\boldsymbol\Sigma_j,o_j,\mathbf c_j)$ denotes its
center, covariance, opacity, and appearance parameters. We call an index
\emph{persistent} if it refers to the same stored scene state independently
of the viewing camera. Thus, even when appearance is view dependent, the
record indexed by $j$ remains identifiable across views. Conventional
language Gaussian fields exploit this property by attaching a normalized
vision--language feature
$\mathbf z_j\in\Sph^{D-1}$ to each stored primitive
\cite{qin2024langsplat,shi2024legaussians,zhou2024feature3dgs}.
For a normalized text embedding $\mathbf e_t\in\Sph^{D-1}$, relevance is
measured by the cosine score
$\langle\mathbf z_j,\mathbf e_t\rangle$.

\paragraph{Anchor-decoded Gaussian scenes.}
We are given posed RGB observations
\[
  \mathcal D
  =
  \{(I_v,\pi_v,\mathbf o_v)\}_{v=1}^{V},
\]
where $\pi_v$ is the calibrated camera projection and $\mathbf o_v$ is the
camera center, together with a frozen anchor-decoded Gaussian
reconstruction. The representation stores persistent anchors
$\mathcal A=\{a_n\}_{n=1}^{N}$. Anchor $a_n$ has position
$\mathbf x_n$ and $K$ indexed decoder outputs. We call the anchor-local index
$k$ a \emph{slot}, so $(n,k)$ is a persistent decoder address.

Let $\boldsymbol\Delta_{nk}$ be the learned anchor-local offset of slot
$(n,k)$ and
$\boldsymbol\mu_{nk}=\mathbf x_n+\boldsymbol\Delta_{nk}$ its canonical
center. Let $D_\theta$ denote the frozen view-conditioned decoder of the
reconstruction, parameterized by $\theta$. Given the stored state of anchor
$a_n$, slot index $k$, and camera--anchor condition $\xi_{vn}$, it predicts
the covariance, opacity, and appearance of the corresponding child Gaussian:
\begin{equation}
  \begin{aligned}
    (\boldsymbol\Sigma_{vnk},o_{vnk},\mathbf c_{vnk})
    &=
    D_\theta(a_n,k,\xi_{vn}), \\
    \widetilde g_{vnk}
    &=
    (\boldsymbol\mu_{nk},
     \boldsymbol\Sigma_{vnk},
     o_{vnk},
     \mathbf c_{vnk}).
  \end{aligned}
  \label{eq:anchor_decoder}
\end{equation}
View- and level-of-detail-dependent selection retains an active subset
$\mathcal K_v(n)\subseteq\{1,\ldots,K\}$, and the rasterizer receives
$
  \mathcal G_v
  =
  \left\{
    \widetilde g_{vnk}
    \,\middle|\,
    1\leq n\leq N,\;
    k\in\mathcal K_v(n)
  \right\}.
$
The slot address $(n,k)$ and its canonical center persist across views, but
the complete child Gaussian is a view-conditioned realization: its covariance,
opacity, appearance, and active state may change with the camera. The child is
therefore not a separately stored semantic owner. Importantly, we do not
assume that a slot has semantics a priori; it only provides a persistent
address to which observations can be assigned.

\paragraph{Rendering responsibility.}
For pixel $p$ in view $v$, let $\alpha_{vnk}(p)\in[0,1]$ be the projected
opacity of an active child. Its front-to-back alpha-compositing contribution is
\begin{equation}
  \omega_{vnk}(p)
  =
  \alpha_{vnk}(p)
  \prod_{(m,\ell)\prec_{v,p}(n,k)}
  \left(1-\alpha_{vm\ell}(p)\right),
  \label{eq:child_weight}
\end{equation}
where $(m,\ell)\prec_{v,p}(n,k)$ denotes a child composited earlier along the
ray through $p$. We set $\omega_{vnk}(p)=0$ when slot $(n,k)$ is inactive.
Thus, $\omega_{vnk}(p)$ measures the contribution of the child generated from
slot $(n,k)$ to the rendered pixel, accounting jointly for its footprint,
opacity, and occlusion.

\paragraph{Persistent language-field objective.}
A fixed 2D vision--language engine provides a unit observation
$\mathbf f_v(p)\in\Sph^{D-1}$ for pixels
$p\in\Omega_v^{\mathrm{obs}}$, where
$\Omega_v^{\mathrm{obs}}$ denotes pixels covered by the 2D engine. We seek
a persistent hierarchical language field
\begin{equation}
  \mathcal Z
  =
  \{\mathbf z_n\}_{n=1}^{N}
  \cup
  \{\mathbf z_{nk}\}_{n=1,k=1}^{N,K},
  \qquad
  \mathbf z_n,\mathbf z_{nk}\in\Sph^{D-1},
  \label{eq:language_field}
\end{equation}
where slots provide fine-grained addresses and anchors provide their coarser
parents. A text query $\mathbf e_t\in\Sph^{D-1}$ scores either level by
$\langle\mathbf z,\mathbf e_t\rangle$.

The reconstruction, decoder, and 2D engine remain frozen throughout field
construction, and no scene-specific semantic labels or feature-field
optimization are used. The challenge is to route observations from
view-conditioned children to persistent slots, obtain a consistent anchor
summary, recover slots with sparse or zero evidence, and avoid storing a full
$D$-dimensional feature at all $NK$ slot addresses.

\section{LangStreet}
\label{sec:language}

\begin{figure*}[t]
  \centering
  \includegraphics[width=0.98\textwidth]{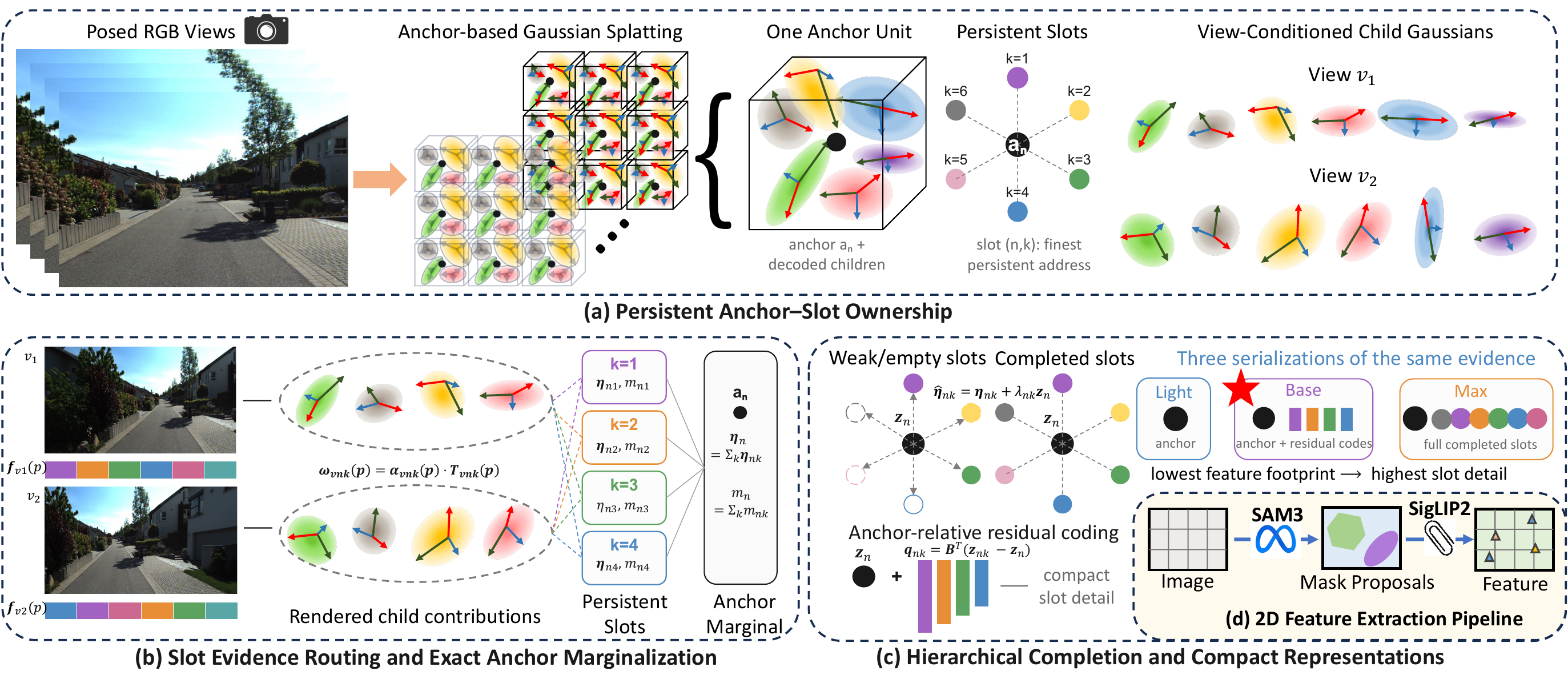}
  \caption{\textbf{LangStreet overview.}
  A fixed SAM~3/SigLIP~2 engine \textbf{(d)} extracts unit language observations from
  posed images. A frozen anchor-decoded reconstruction \textbf{(a)} retains persistent
  anchor--slot addresses while instantiating view-conditioned child Gaussians.
  Their compositing responsibilities route 2D observations to additive slot
  evidence, whose unnormalized statistics sum exactly to the anchor marginal
  \textbf{(b)}. Anchor-aligned completion recovers weakly supported slots, and
  anchor-relative residual coding yields \ours~(base), with anchor-only
  \ours~(light) and full-slot \ours~(max) providing compact and high-capacity
  variants, respectively \textbf{(c)}.}
  \label{fig:method}
\end{figure*}

Figure~\ref{fig:method} summarizes \ours. Section~\ref{sec:responsibility}
routes 2D observations to persistent slots and marginalizes them to anchors;
Sec.~\ref{sec:twolevel} completes weakly supported slots while preserving the
anchor direction; Sec.~\ref{sec:residual_coding} derives the compact
\ours~(base) representation; and Sec.~\ref{sec:implementation} describes
construction and query readout. The reconstruction, decoder, and 2D engine
remain frozen throughout.

\subsection{Renderer-Routed Additive Evidence}
\label{sec:responsibility}
\label{sec:spherical}

As illustrated in Fig.~\ref{fig:method}(b), the frozen renderer already
specifies how each view-conditioned child contributes to a pixel. For
$p\in\Omega_v^{\mathrm{obs}}$, define its preceding transmittance as
\begin{equation}
  T_{vnk}(p)
  \coloneqq
  \prod_{(m,\ell)\prec_{v,p}(n,k)}
  \left(1-\alpha_{vm\ell}(p)\right).
  \label{eq:child_transmittance}
\end{equation}
The corresponding child responsibility is
$\omega_{vnk}(p)=\alpha_{vnk}(p)T_{vnk}(p)$, with
$\omega_{vnk}(p)=0$ when slot $(n,k)$ is inactive. This renderer-native weight
routes language while already accounting for projected footprint, opacity,
and occlusion.

We accumulate the weighted language observations and their support at each
persistent slot, then sum the same statistics to its parent anchor:
\begin{equation}
  \begin{alignedat}{2}
    \boldsymbol{\eta}_{nk}
    &\coloneqq
    \sum_{\substack{v=1,\ldots,V\\
                    p\in\Omega_v^{\mathrm{obs}}}}
    \omega_{vnk}(p)\,\mathbf f_v(p),
    &\quad
    \boldsymbol{\eta}_{n}
    &\coloneqq
    \sum_{k=1}^{K}\boldsymbol{\eta}_{nk},
    \\[-1pt]
    m_{nk}
    &\coloneqq
    \sum_{\substack{v=1,\ldots,V\\
                    p\in\Omega_v^{\mathrm{obs}}}}
    \omega_{vnk}(p),
    &\quad
    m_n
    &\coloneqq
    \sum_{k=1}^{K}m_{nk}.
  \end{alignedat}
  \label{eq:additive_evidence}
\end{equation}
Here $\boldsymbol\eta_{nk}$ is unnormalized directional evidence and $m_{nk}$
is its observation support. Because these statistics are additive, the
right-hand definitions are exact anchor marginals; normalized feature
directions would not satisfy this property.

For nonzero resultants, the raw slot and anchor directions are
\begin{equation}
  \mathbf z^{\mathrm{raw}}_{nk}
  =
  \frac{\boldsymbol\eta_{nk}}{\|\boldsymbol\eta_{nk}\|_2},
  \qquad
  \mathbf z_n
  =
  \frac{\boldsymbol\eta_n}{\|\boldsymbol\eta_n\|_2}.
  \label{eq:raw_directions}
\end{equation}
A node with resultant norm at most $\epsilon$ has no valid direction and
abstains; its mass remains available as a support statistic.

\subsection{Direction-Preserving Hierarchical Completion}
\label{sec:twolevel}

Slots preserve sub-anchor detail but receive uneven evidence, whereas their
anchor is better supported and spatially coarser. We therefore complete weak
slots in evidence space. For any nonnegative anchor-aligned mass
$\lambda_{nk}$, define
\begin{equation}
  \widehat{\boldsymbol\eta}_{nk}
  \coloneqq
  \boldsymbol\eta_{nk}+\lambda_{nk}\mathbf z_n,
  \qquad
  \mathbf z^{\mathrm{pool}}_{nk}
  =
  \frac{\widehat{\boldsymbol\eta}_{nk}}
       {\|\widehat{\boldsymbol\eta}_{nk}\|_2}.
  \label{eq:anchor_completion}
\end{equation}
The second expression is used only when its numerator is nonzero; an anchor
without a valid direction contributes no completion mass.

\paragraph{Anchor-direction preservation.}
Let $\Lambda_n=\sum_k\lambda_{nk}$. Since
$\mathbf z_n=\boldsymbol\eta_n/\|\boldsymbol\eta_n\|_2$,
\begin{equation}
  \sum_{k=1}^{K}\widehat{\boldsymbol\eta}_{nk}
  =
  \left(
    1+\frac{\Lambda_n}{\|\boldsymbol\eta_n\|_2}
  \right)\boldsymbol\eta_n.
  \label{eq:marginal_preservation}
\end{equation}
Thus, any nonnegative anchor-aligned completion changes the anchor resultant
only by a positive scale and preserves its normalized direction.

We allocate completion according to sibling agreement and target-slot support:
\begin{equation}
  R_n
  \coloneqq
  \frac{\|\boldsymbol\eta_n\|_2}
       {\max\!\left(\sum_{k=1}^{K}\|\boldsymbol\eta_{nk}\|_2,\epsilon\right)},
  \qquad
  \bar m_n
  \coloneqq
  \frac{m_n}{K},
  \label{eq:sibling_agreement}
\end{equation}
\begin{equation}
  \lambda_{nk}
  =
  \beta R_n\bar m_n
  \frac{\bar m_n}{\bar m_n+m_{nk}},
  \qquad
  \beta\geq0.
  \label{eq:support_pooling}
\end{equation}
By the triangle inequality, $R_n\in[0,1]$ and measures agreement among sibling
slot resultants. The remaining factors set an anchor-specific evidence scale
and concentrate completion on poorly supported slots. Hence an unsupported
slot inherits the anchor direction, while the update becomes negligible as
its own evidence grows.

\subsection{Anchor-Relative Coding}
\label{sec:residual_coding}

A full $D$-dimensional direction at every completed slot repeatedly stores
content already shared with its anchor. We instead use the anchor as a local
semantic origin and define
\begin{equation}
  \boldsymbol\delta_{nk}
  \coloneqq
  \mathbf z^{\mathrm{pool}}_{nk}-\mathbf z_n.
  \label{eq:anchor_residual}
\end{equation}
We fit a rank-$r$ orthonormal basis $B\in\mathbb R^{D\times r}$ to supported
residuals using truncated SVD. Each slot stores only
\begin{equation}
  \mathbf q_{nk}
  =
  B^\top\boldsymbol\delta_{nk},
  ~~
  \widetilde{\mathbf z}_{nk}
  =
  \operatorname{normalize}\!\left(\mathbf z_n+B\mathbf q_{nk}\right).
  \label{eq:residual_decode}
\end{equation}
The shared basis captures recurring within-anchor variation, while
$\mathbf q_{nk}$ retains slot-specific detail. This coding step uses no
semantic labels or feature-field optimization.

The same construction yields three storage variants:
\begin{equation}
  \mathbf z^{\mathrm{light}}_n=\mathbf z_n,
  \qquad
  \mathbf z^{\mathrm{base}}_{nk}=\widetilde{\mathbf z}_{nk},
  \qquad
  \mathbf z^{\mathrm{max}}_{nk}=\mathbf z^{\mathrm{pool}}_{nk}.
  \label{eq:three_tiers}
\end{equation}
We use base as our primary configuration. Light omits the slot codes, whereas
max stores every completed slot direction explicitly. Ignoring lower-order
support and validity metadata, their dominant directional payloads are
\begin{equation}
  \mathcal C_{\mathrm{light}}=ND,
  \ 
  \mathcal C_{\mathrm{base}}=ND+NKr+Dr,
  \ 
  \mathcal C_{\mathrm{max}}=NKD.
  \label{eq:tier_storage}
\end{equation}
Since $r\ll D$, base retains slot-level variation with substantially less
storage than max.

\subsection{Construction and Query Readout}
\label{sec:implementation}
\label{sec:querying}

\paragraph{Closed-form construction.}
Let $\mathcal R_v(\{\mathbf h_{vnk}\})$ denote alpha-compositing of arbitrary
per-child channels. Linearity in $\mathbf h_{vnk}$ gives
\begin{equation}
  \nabla_{\mathbf h_{vnk}}
  \sum_{p\in\Omega_v^{\mathrm{obs}}}
  \left\langle
    \mathcal R_v(\{\mathbf h\})(p),\mathbf f_v(p)
  \right\rangle
  =
  \sum_{p\in\Omega_v^{\mathrm{obs}}}
  \omega_{vnk}(p)\mathbf f_v(p).
  \label{eq:backward_lifting}
\end{equation}
A standard renderer backward pass therefore accumulates
$\boldsymbol\eta_{nk}$; a scalar seed of ones analogously accumulates
$m_{nk}$. We use the stock \texttt{gsplat} rasterizer
\cite{ye2025gsplat}, process feature channels in blocks, and apply completion
and coding after accumulation, without a semantic CUDA kernel or scene-specific
semantic optimization.

\paragraph{Text scoring and readout.}
For any stored or decoded unit direction $\mathbf z$ and normalized text
embedding $\mathbf e_t$, we use the cosine score
$s_t(\mathbf z)\coloneqq\langle\mathbf z,\mathbf e_t\rangle$.
In light, each active child inherits its anchor direction; in base and max it
inherits the corresponding decoded or explicit slot direction. A rendered 2D
query composites these scores with the same responsibilities:
\begin{equation}
  \begin{aligned}
    S^\tau_{v,t}(p)
    &=
    \sum_{n=1}^{N}
    \sum_{k\in\mathcal K_v(n)}
    \omega_{vnk}(p)\,
    s_t\!\left(\mathbf z^\tau_{nk}\right),
    \\[-1pt]
    \tau
    &\in
    \{\mathrm{light},\mathrm{base},\mathrm{max}\}.
  \end{aligned}
  \label{eq:rendered_query}
\end{equation}
where $\mathbf z^{\mathrm{light}}_{nk}\coloneqq\mathbf z_n$. A 3D query
instead accesses a declared persistent anchor or canonical slot address; the
spatial assignment and fallback rule are specified with each evaluation.
These address-selection rules are readout policies, not part of the stored
field. Replacing the text bank changes only the dot products and does not
rebuild the scene or language field.

\section{Experiments}
\label{sec:experiments}

\subsection{Datasets}
\label{sec:datasets}

\paragraph{KITTI-360.}
We evaluate eight independently reconstructed \kitti street windows
\cite{liao2023kitti360}. Official validation labels define the 2D protocol,
and accumulated semantic points define the 3D protocol under the same
17-class mapping; unsupported points count as errors.

\paragraph{Virtual KITTI 2.}
We use all five \vkitti clones \cite{cabon2020virtualkitti2}. Camera~0 provides
construction observations, while camera~1 is held out for 2D evaluation and
depth-backprojected 3D evaluation. Released depth only seeds reconstruction,
making \vkitti a controlled diagnostic of sparse slot evidence.

\paragraph{Waymo.}
We evaluate an eleven-segment core from \waymo's five-camera rig
\cite{sun2020waymo} as an independent benchmark. We report matched 2D results,
but no 3D headline because it lacks a semantic-point target comparable to
\kitti.

\begin{table*}[htbp]
  \centering
  \mainwidefont
  \renewcommand{\arraystretch}{1.10}
  \setlength{\tabcolsep}{2.75pt}
  \resizebox{\fulltablewidth}{!}{%
  \begin{tabular}{@{}c@{\hspace{1.05em}}c@{}}
    \textbf{(a) KITTI-360} & \textbf{(b) Virtual KITTI 2}\\[-0.15em]
    \begin{tabular}{@{}l c c c c c c@{}}
      \toprule
      Method & \psnr\up & 2D \up & 3D \up & Build \down & Payload \down & FPS \up\\
      \midrule
      Image Teacher (SAM3) & --- & 32.91 & --- & --- & --- & ---\\
\midrule
LangSplat (anchor) & 20.28 & 30.26 & 12.28 & 506.6\,s & \textcolor{lsred}{\textbf{1.29\,GiB}} & \textcolor{lsred}{\textbf{861}}\\
Feature 3DGS & 20.28 & \textcolor{lsred}{\textbf{34.20}} & 16.29 & 124.0\,s & 12.90\,GiB & 137\\
OccamLGS & 20.28 & 33.33 & 15.28 & \textcolor{lsred}{\textbf{13.3\,s}} & 12.90\,GiB & 137\\
LUDVIG & 20.28 & 33.27 & 16.27 & 235.0\,s & 12.90\,GiB & 137\\
SFS & 20.28 & 33.64 & 13.67 & 153.9\,min & 12.90\,GiB & 137\\
VALA & 20.28 & 33.64 & 15.75 & 95.3\,min & 12.90\,GiB & 137\\
\midrule
\ours~(light) & 20.28 & 33.70 & 17.63 & \textcolor{lsblue}{\underline{95.3\,s}} & \textcolor{lsred}{\textbf{1.29\,GiB}} & \textcolor{lsblue}{\underline{860}}\\
\textbf{\ours~(base)} & 20.28 & \textcolor{lsblue}{\underline{34.19}} & \textcolor{lsred}{\textbf{20.62}} & 22.3\,min & \textcolor{lsblue}{\underline{2.72\,GiB}} & 145\\
\ours~(max) & 20.28 & \textcolor{lsred}{\textbf{34.20}} & \textcolor{lsblue}{\underline{20.56}} & 154.6\,s & 12.90\,GiB & 137\\

      \bottomrule
    \end{tabular}
    &
    \begin{tabular}{@{}l c c c c c c@{}}
      \toprule
      Method & \psnr\up & 2D \up & 3D \up & Build \down & Payload \down & FPS \up\\
      \midrule
      Image Teacher (SAM3) & --- & 42.69 & --- & --- & --- & ---\\
\midrule
LangSplat (anchor) & 20.94 & 31.82 & 10.59 & 12.6\,min & \textcolor{lsred}{\textbf{0.57\,GiB}} & \textcolor{lsred}{\textbf{1261}}\\
Feature 3DGS & 20.94 & \textcolor{lsblue}{\underline{42.86}} & 22.32 & \textcolor{lsblue}{\underline{153.2\,s}} & 5.69\,GiB & 372\\
OccamLGS & 20.94 & 41.78 & 22.10 & \textcolor{lsred}{\textbf{17.5\,s}} & 5.69\,GiB & 371\\
LUDVIG & 20.94 & 41.39 & 22.71 & 174.8\,s & 5.69\,GiB & 372\\
SFS & 20.94 & 42.67 & 18.38 & 187.3\,min & 5.69\,GiB & 372\\
VALA & 20.94 & 42.37 & 21.96 & 95.3\,min & 5.69\,GiB & 371\\
\midrule
\ours~(light) & 20.94 & 42.58 & 22.70 & 238.6\,s & \textcolor{lsred}{\textbf{0.57\,GiB}} & \textcolor{lsblue}{\underline{1258}}\\
\textbf{\ours~(base)} & 20.94 & \textcolor{lsred}{\textbf{42.87}} & \textcolor{lsblue}{\underline{25.25}} & 281.0\,s & \textcolor{lsblue}{\underline{1.20\,GiB}} & 372\\
\ours~(max) & 20.94 & \textcolor{lsred}{\textbf{42.87}} & \textcolor{lsred}{\textbf{25.30}} & 165.4\,s & 5.69\,GiB & 371\\

      \bottomrule
    \end{tabular}
  \end{tabular}}
  \caption{\textbf{Controlled comparison on \kitti and \vkitti.}
  All methods share reconstruction, observations, prompts, renderer, and
  evaluator. \textcolor{lsred}{\textbf{Red bold}} and
  \textcolor{lsblue}{\underline{blue underlined}} mark the best and
  second-best comparable results.}
  \label{tab:main_kitti_vkitti}
  \label{tab:main_baselines}
  \label{tab:method_ablation}
\end{table*}

\begin{table}[!t]
  \centering
  \mainsinglefont
  \renewcommand{\arraystretch}{1.08}
  \setlength{\tabcolsep}{2.55pt}
  \resizebox{\columnwidth}{!}{%
  \begin{tabular}{@{}l c c c c c@{}}
    \toprule
    Method & \psnr\up & 2D \up & Build \down & Payload \down & FPS \up\\
    \midrule
    Image Teacher (SAM3) & --- & 15.55 & --- & --- & ---\\
\midrule
LangSplat (anchor) & 27.67 & 7.95 & 17.0\,min & \textcolor{lsred}{\textbf{1.10\,GiB}} & \textcolor{lsred}{\textbf{872}}\\
Feature 3DGS & 27.67 & \textcolor{lsred}{\textbf{16.27}} & 147.9\,s & 11.04\,GiB & 144\\
OccamLGS & 27.67 & 15.39 & \textcolor{lsred}{\textbf{140.8\,s}} & 11.04\,GiB & 144\\
SFS & 27.67 & 14.06 & 349.6\,min & 11.04\,GiB & 144\\
VALA & 27.67 & 16.06 & 180.1\,min & 11.04\,GiB & 144\\
\midrule
\ours~(light) & 27.67 & 16.14 & \textcolor{lsblue}{\underline{147.8\,s}} & \textcolor{lsred}{\textbf{1.10\,GiB}} & \textcolor{lsblue}{\underline{866}}\\
\textbf{\ours~(base)} & 27.67 & \textcolor{lsblue}{\underline{16.25}} & 795.5\,s & \textcolor{lsblue}{\underline{2.32\,GiB}} & 144\\
\ours~(max) & 27.67 & \textcolor{lsred}{\textbf{16.27}} & 168.3\,s & 11.04\,GiB & 144\\

    \bottomrule
  \end{tabular}}
  \caption{\textbf{Controlled comparison on \waymo.}
  All rows use the same eleven-segment evaluation set and the same incremental
  field-construction accounting.}
  \label{tab:main_waymo}
\end{table}

\subsection{Implementation Details}
\label{sec:implementation_details}

All experiments run on one NVIDIA H200. The 3D reconstructions and per-view
SAM~3/SigLIP~2 observations are preprocessed once and shared by every method.
We accumulate evidence in FP32, store semantic payloads in FP16, and use
rank-$128$ codes for \ours~(base) unless stated otherwise. Reconstruction,
decoder, and 2D language engine remain frozen during field construction.

\subsection{Benchmarking Protocol}
\label{sec:protocol}
\label{sec:baselines}

\paragraph{Reconstruction and baselines.}
Each segment is reconstructed with an Octree-GS anchor decoder
\cite{ren2025octree} using the \texttt{gsplat} MCMC recipe
\cite{ye2025gsplat}, then frozen. We compare controlled adaptations of
LangSplat, Feature 3DGS, Occam's LGS, LUDVIG, Splat Feature Solver, and VALA
\cite{qin2024langsplat,zhou2024feature3dgs,cheng2025occam,
marrie2025ludvigofficial,xiong2026sfs,wang2025vala}. They share geometry,
views, observations, prompts, renderer, and evaluator, so the comparison
isolates language-field estimation rather than reproducing each original
end-to-end system. LangSplat is anchor-adapted because native slot optimization
exceeds the memory of a single H200. Image Teacher accesses the target image
and is a privileged reference, not a persistent 3D field.

\paragraph{Evaluation and accounting.}
We report standard 2D and 3D mIoU; \psnr only verifies the shared
reconstruction. Each representation uses its declared persistent readout, with
a common-readout control in the supplement. \emph{Build} measures incremental
field construction after reconstruction and 2D preprocessing, including
lifting, completion or solving, coding, and serialization. \emph{Payload}
counts stored FP16 semantic variables and excludes shared geometry and 2D
features. \emph{FPS} measures scoring and rendering for a pre-encoded text
embedding. Peak memory and scratch cover incremental field construction and
exclude shared inputs; detailed traces are in the supplement.

\subsection{Quantitative Results}
\label{sec:main_results}

\paragraph{Base balances accuracy and storage.}
Tables~\ref{tab:main_kitti_vkitti} and~\ref{tab:main_waymo} show that base
nearly matches max while using far less semantic storage. On \kitti, it is
within $0.01$ 2D mIoU of max using $2.72$ rather than $12.90$\,GiB; on
\waymo it remains within $0.02$ mIoU at a $4.8\times$ smaller payload. Light
is cheaper but loses fine spatial detail. Anchor-relative coordinates therefore
make slot detail inexpensive enough to retain.

\begin{figure*}[!t]
  \centering
  \includegraphics[width=\textwidth]{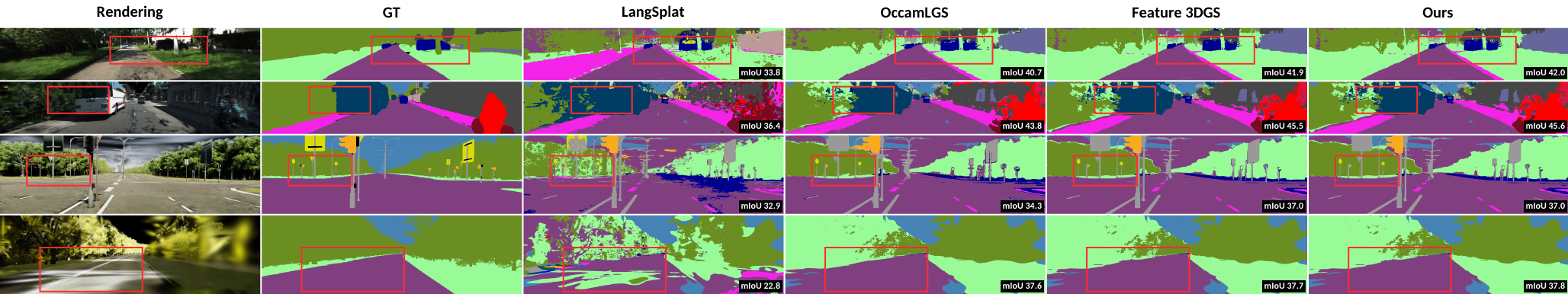}
  \caption{\textbf{Qualitative segmentation results on \kitti and \vkitti.}
  We compare RGB, ground truth, matched baselines, and \ours~(base) on
  representative frames.}
  \label{fig:quali_kitti_vkitti}
\end{figure*}

\begin{figure*}[!t]
  \centering
  \includegraphics[width=\textwidth]{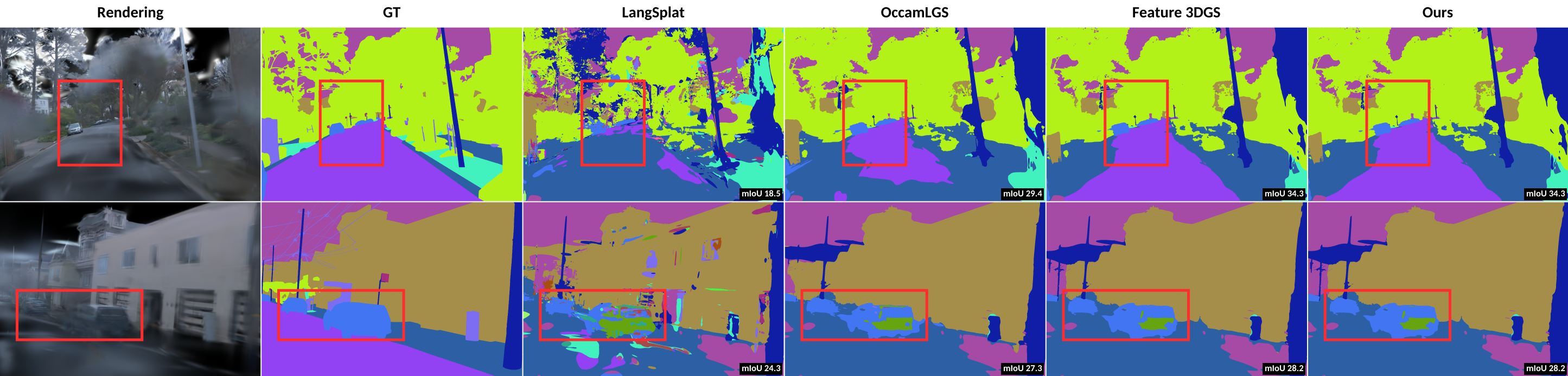}
  \caption{\textbf{Qualitative segmentation results on \waymo.}
  We compare RGB, ground truth, matched baselines, and \ours~(base) on
  representative frames.}
  \label{fig:quali_waymo}
\end{figure*}

\paragraph{Completion improves spatial access, not image fitting.}
Under the fixed hierarchical readout in Tab.~\ref{tab:ablations}(b),
support-aware completion raises 3D mIoU from $18.47$ to $20.56$ on
\kitti and from $23.12$ to $25.30$ on \vkitti, while leaving 2D accuracy
essentially unchanged. The canonical-slot control in the supplement shows
that the field-level gain is substantially larger under sparse \vkitti
support. Completion therefore repairs weak persistent addresses rather than
fitting construction images more aggressively.

\paragraph{Representation and estimator are complementary.}
Direct lifting, diffusion, robust aggregation, and coupled solving trade
accuracy against construction cost, but do not resolve where semantics should
persist or how dense slot detail should be stored. \ours~(base) addresses this
orthogonal representation problem, giving the strongest \kitti 3D result and
near-max accuracy on all three datasets. The controlled study is not an
official ranking of the original systems.

\subsection{Qualitative Results}
\label{sec:qualitative}

\begin{figure}[!t]
  \centering
  \includegraphics[width=\columnwidth]{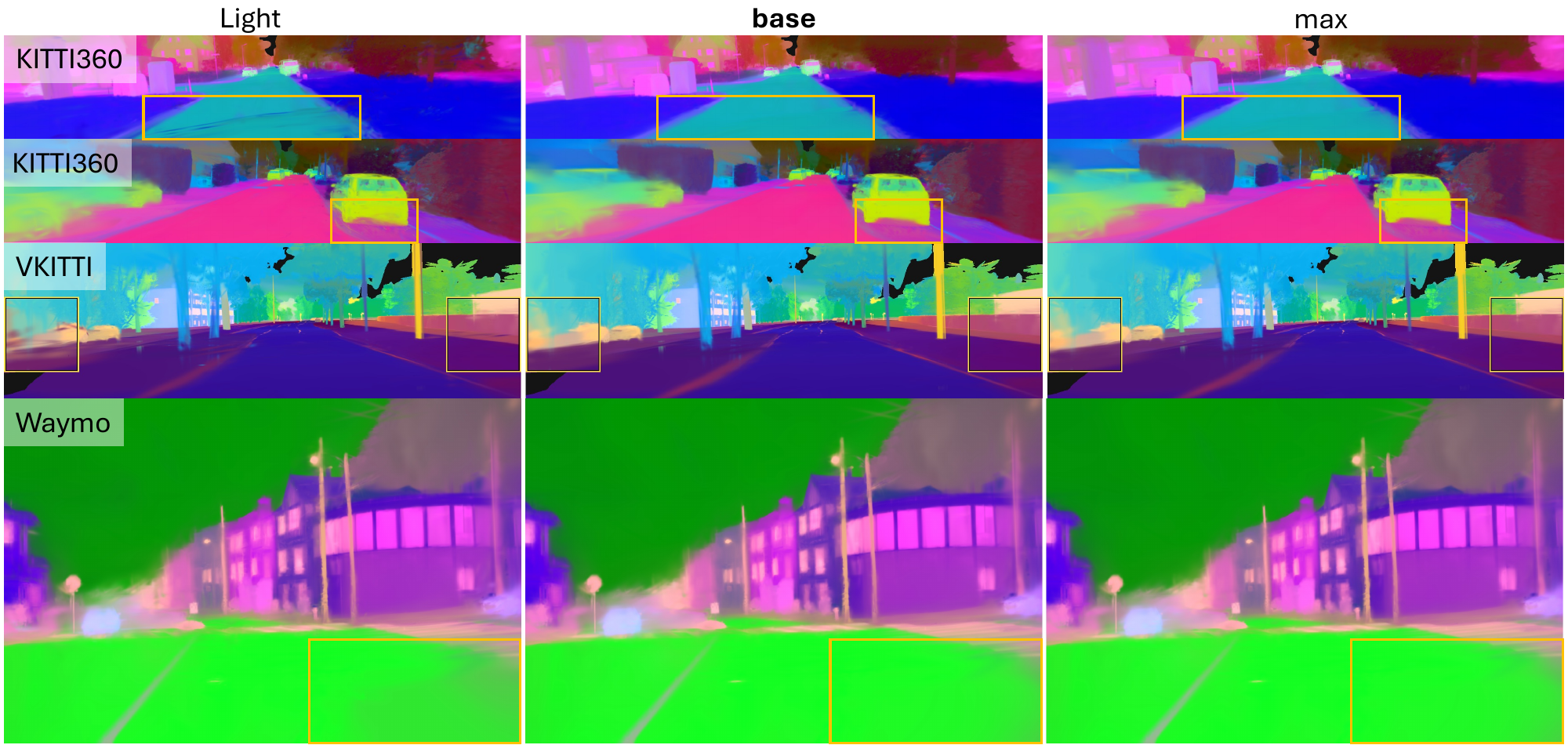}
  \caption{\textbf{PCA visualizations of light, base, and max features.}
  A shared projection within each scene shows that base preserves the
  fine-grained structure of max, while light exhibits anchor-level smoothing.}
  \label{fig:tier_feature_pca}
\end{figure}

Figures~\ref{fig:quali_kitti_vkitti}--\ref{fig:tier_feature_pca} follow the
same rate--distortion trend. Base and max preserve thin structures and object
boundaries that light smooths; completion fills sparse slot holes, while
mixed-surface anchors remain the characteristic failure case. The same
ordering also holds on \waymo.

\section{Ablation Study}
\label{sec:analysis}

Table~\ref{tab:ablations} follows the three method stages. Block~(a) traces the
progressive deployed ladder; block~(b) fixes full-dimensional slots and one
hierarchical readout while varying completion; block~(c) fixes that completed
field and readout while varying only its code.

\begin{table}[!t]
  \centering
  \mainwidefont
  \renewcommand{\arraystretch}{1.04}
  \setlength{\tabcolsep}{2.75pt}
  \resizebox{.45\textwidth}{!}{%
  \begin{tabular}{@{}l c c c c c c@{}}
    \toprule
    & \multicolumn{3}{c}{\kitti} & \multicolumn{3}{c}{\vkitti}\\
    \cmidrule(lr){2-4}\cmidrule(lr){5-7}
    Variant & 2D \up & 3D \up & GiB \down & 2D \up & 3D \up & GiB \down\\
    \midrule
\multicolumn{7}{@{}l}{\emph{(a) Progressive hierarchy and operating points (declared readouts)}}\\
Raw slot field (no anchor sharing) & 34.20 & 16.29 & 12.90 & 42.86 & 22.32 & 5.69\\
$+$ exact anchor marginal $=$ \ours~(light) & 33.70 & 17.63 & 1.29 & 42.58 & 22.70 & 0.57\\
$+$ support-aware completion $=$ \ours~(max) & 34.20 & 20.56 & 12.90 & 42.87 & 25.30 & 5.69\\
$+$ rank-128 residual coding $=$ \textbf{\ours~(base)} & 34.19 & 20.62 & 2.72 & 42.87 & 25.25 & 1.20\\
\midrule
\multicolumn{7}{@{}l}{\emph{(b) Completion under the fixed deployed hierarchical readout}}\\
No completion ($\lambda_{nk}{=}0$) & 34.20 & 18.47 & 12.90 & 42.86 & 23.12 & 5.69\\
Empty-slot fallback & 34.20 & 19.59 & 12.90 & 42.86 & 22.81 & 5.69\\
Uniform pseudo-mass & 34.11 & 20.01 & 12.90 & 42.83 & 22.96 & 5.69\\
Leave-one-out anchor prior & 34.14 & 20.54 & 12.90 & 42.86 & 25.30 & 5.69\\
Support-aware, Eq.~\eqref{eq:support_pooling} $=$ \ours~(max field) & 34.20 & 20.56 & 12.90 & 42.87 & 25.30 & 5.69\\
\midrule
\multicolumn{7}{@{}l}{\emph{(c) Coding the same completed field under the same readout}}\\
Full completed slots $=$ \ours~(max) & 34.20 & 20.56 & 12.90 & 42.87 & 25.30 & 5.69\\
Anchor-relative, rank 128 $=$ \textbf{\ours~(base)} & 34.19 & 20.62 & 2.72 & 42.87 & 25.25 & 1.20\\
Anchor-relative, rank 8 & 33.53 & 19.88 & 1.38 & 42.80 & 25.06 & 0.61\\

    \bottomrule
  \end{tabular}}
  \caption{\textbf{Progressive and controlled ablations.}
  (a)~follows the deployed $+$ ladder with each tier's declared readout.
  (b)~fixes full slots and one hierarchical readout, varying only completion;
  its final row is max and the source field for base. (c)~fixes this field and
  readout, varying only coding.}
  \label{tab:ablations}
  \label{tab:component_ablation}
  \label{tab:support_shrinkage}
  \label{tab:coding}
\end{table}

\paragraph{(a) Why use the native hierarchy?}
The continuous $+$ ladder maps each operation to a model: exact anchor
marginalization gives light, support-aware completed slots give max, and
anchor-relative coding turns the same slot field into base. Block~(a) is
therefore a system progression that reproduces the LangStreet rows in
Tab.~\ref{tab:main_kitti_vkitti}; blocks~(b) and~(c) provide controlled
attribution.

\paragraph{(b) Why support-aware completion?}
All rows use the same full slot capacity and hierarchical readout, so only
$\lambda_{nk}$ changes. Support-aware completion improves 3D mIoU from $18.47$
to $20.56$ on \kitti and from $23.12$ to $25.30$ on \vkitti while retaining
2D accuracy. On \kitti, increasingly evidence-aware priors improve steadily;
on \vkitti, empty-slot fallback and uniform mass degrade the raw field, whereas
sibling-conditioned priors recover sparse addresses. Leave-one-out reaches the
same \vkitti score, but constructs a different prior for every slot; our rule
slightly improves \kitti and 2D accuracy while retaining one shared anchor
prior and the direction-preservation guarantee in
Eq.~\eqref{eq:marginal_preservation}. Its final row is the max field and the
field coded to form base.

\paragraph{(c) Why code relative to anchors?}
With field and readout fixed, rank-$128$ stays within $0.01$ 2D mIoU of max on
\kitti while reducing payload from $12.90$ to $2.72$\,GiB; on \vkitti it
reduces payload from $5.69$ to $1.20$\,GiB. Rank-$8$ nearly halves the coded
payload again, but lowers \kitti 2D/3D mIoU from $34.20/20.56$ to
$33.53/19.88$; the loss is smaller on \vkitti, from $42.87/25.30$ to
$42.80/25.06$. We therefore use rank-$128$ for the default base model and
report rank-$8$ as a more aggressive compression setting.

\paragraph{Completion versus readout.}
Using the deployed readout in block~(b) remains controlled because it is
identical for every completion rule. The complementary canonical-slot test in
the supplement attributes $0.31$ and $2.18$ 3D points to completion on
\kitti and \vkitti, respectively; a readout-only test measures the additional
gain from hierarchical fallback. Thus the raw-to-max progression in block~(a)
combines two distinct effects: completion repairs weak semantic evidence,
whereas fallback improves coverage of persistent spatial addresses.

\begin{figure}[!t]
  \centering
  \includegraphics[width=\columnwidth]{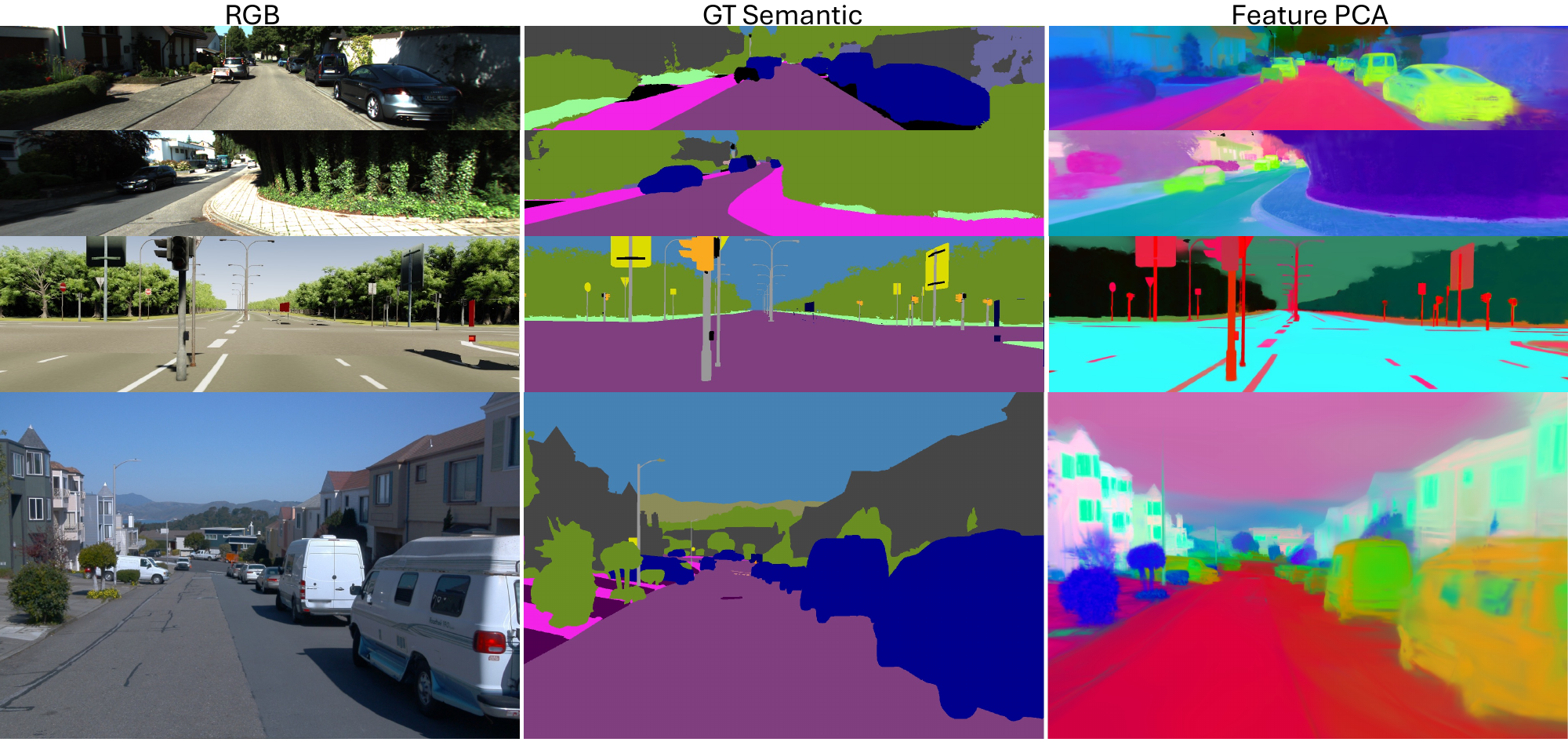}
  \caption{\textbf{Granularity beyond the benchmark taxonomy.}
  The 17-class mask merges distinct surfaces, while a label-free PCA projection
  retains finer local structure.}
  \label{fig:feature_granularity}
\end{figure}

\begin{table}[!t]
  \centering
  \mainsinglefont
  \renewcommand{\arraystretch}{1.06}
  \setlength{\tabcolsep}{2.75pt}
  \resizebox{\columnwidth}{!}{%
  \begin{tabular}{@{}l l c c@{}}
    \toprule
    Teacher & Added oracle & 2D \up & slot 3D \up\\
    \midrule
Automatic priors (SAM~3 $+$ SigLIP~2) & - & 42.87 & 25.25\\
$+$ GT regions (encoder kept) & $+$ regions & 47.26 & 29.65\\
$+$ GT regions $+$ GT labels & $+$ labels & 68.85 & 39.36\\

    \bottomrule
  \end{tabular}}
  \caption{\textbf{Observation ceiling on \vkitti.}
  Reconstruction and field estimation are fixed; only observations change.}
  \label{tab:prior_gt}
\end{table}

\FloatBarrier
\paragraph{What does fixed-taxonomy mIoU miss?}
Figure~\ref{fig:feature_granularity} shows that moderate mIoU need not imply a
coarse field. Evaluation projects continuous features onto 17 text embeddings
and rewards only the resulting hard label, giving no credit to coherent
within-class material, boundary, or surface variation. PCA exposes this
retained granularity but remains a diagnostic, not a replacement for mIoU.

\paragraph{Where does the remaining ceiling come from?}
In Tab.~\ref{tab:prior_gt}, ground-truth regions improve 2D and slot-3D mIoU by
$4.39$ and $4.40$ points, so observation coverage transfers consistently
through our accumulation. Adding ground-truth labels gives a further $21.59$
points in 2D but $9.71$ in 3D: label alignment directly fixes image decisions,
whereas a 3D address must also be reconstructed, observed, and matched to an
evaluation point. Upstream semantics, geometry, and sparse support therefore
remain distinct bottlenecks.

\paragraph{What survives compact coding?}
Figures~\ref{fig:tier_feature_pca} and~\ref{fig:feature_granularity}, together
with Tab.~\ref{tab:ablations}, show that base preserves coherent within-class
boundaries instead of collapsing features onto the benchmark taxonomy. Light
smooths local variation, while rank-$8$ exposes the expected loss from an
aggressive bitrate. This visual ordering supports the quantitative conclusion
that anchor-relative coding preserves max-level detail until the residual
budget becomes very small.

\paragraph{Cross-dataset consistency.}
The three benchmarks expose complementary regimes. On well-observed \kitti,
coding determines the practical accuracy--storage trade-off. The held-out
camera of \vkitti makes sparse-support completion substantially more visible.
Independently reconstructed \waymo segments preserve the base--max ordering
under a different five-camera rig. These trends indicate that ownership,
completion, and coding address representation-level failures rather than one
dataset-specific artifact.

Additional per-unit results, common-readout and readout-only controls,
support-stratified completion, coding, robustness, efficiency, and qualitative
analyses are provided in the supplement.

\FloatBarrier
\section{Conclusion}
\label{sec:conclusion}

LangStreet makes anchor-decoded street Gaussians language-queryable by
separating transient routing from persistent semantic ownership. Renderer
responsibilities accumulate evidence at stable slots, exact marginalization
recovers anchor summaries, anchor-aligned completion stabilizes sparse support,
and anchor-relative codes retain local detail compactly. Across three driving
benchmarks, base nearly matches max while using substantially less feature
storage, and completion helps most when slot evidence is sparse. More broadly,
scalable language fields should inherit the persistent hierarchy of the
reconstruction and report the stored field separately from its readout.

\phantomsection
\label{page:last_main}
\clearpage

{
    \small
    \bibliographystyle{ieeenat_fullname}
    \bibliography{main,recent,datasets,street,language_fields}
}

\end{document}